\documentclass[a4paper]{article}
\usepackage{amsmath}
\usepackage{amsfonts}
\usepackage{amssymb}
\usepackage{amsthm}
\usepackage[dvips]{graphicx}
\usepackage{graphicx}
\usepackage{color}
\usepackage{epsfig}
\usepackage{floatflt}
\usepackage{tabularx}
\usepackage{eqnarray}
\usepackage{verbatim}
\usepackage{url}
\usepackage{subfigure}
\usepackage{pst-plot}
\usepackage{pstricks-add}
\usepackage{RR}
\usepackage{hyperref}

\def\vect#1{\mbox{\boldmath $#1$}}
\def\x{\vect{x}}

\def\a{{\mathbf a}}

\def\alphab{\mbox{\boldmath$\alpha$}}
\def\thetab{\mbox{\boldmath$\theta$}}

\def\w{{\mathbf w}}
\def\D{{\mathbf D}}

\def\W{{\mathbf W}}
\def\d{{\mathbf d}}

\def\C{{\mathcal C}}

\def\R{{\mathcal R}}
\def\S{{\mathcal S}}
\def\H{{\mathcal H}}
\def\d{{\mathbf d}}

\def\b{{\mathbf b}}

\def\Real{{\mathbb R}}
\def\r{{\mathbf r}}
\def\U{{\mathbf U}}

\def\A{{\mathbf A}}

\def\argmin{\operatornamewithlimits{arg\,min}}

\RRdate{September 2008}

\RRauthor{
Julien Mairal\thanks[inria]{INRIA}%
  \thanks[willow]{WILLOW project-team, Laboratoire d'Informatique de
l'Ecole Normale Sup\'erieure, ENS/INRIA/CNRS UMR 8548}%
  \and
Francis Bach\thanksref{inria}\thanksref{willow}
\and Jean Ponce\thanks{Ecole Normale Sup\'erieure}\thanksref{willow}
\and Guillermo Sapiro\thanks{University of Minnesota, Department of Electrical Engineering}
\and Andrew Zisserman\thanks{University of Oxford}\thanksref{willow}
}
\authorhead{Mairal, Bach, Ponce, Sapiro \& Zisserman}
\RRtitle{Apprentissage de dictionnaires supervis\'e}
\RRetitle{Supervised Dictionary Learning}
\titlehead{Supervised Dictionary Learning}
\RRresume{Il est maintenant bien \'etabli que les repr\'esentations parcimonieuses de signaux sont bien adapt\'ees \`a des taches de restauration d'image, de sons ou de video. De recherches r\'ecentes ont eu pour but d'apprendre des repr\'esentations discriminantes au lieu de seulement reconstructives. Ce travail propose un nouveau cadre pour repr\'esenter des signaux appartenant \`a plusieurs classes diff\'erentes, en apprenant de fa\c{c}on simultan\'ee un dictionnaire partag\'e et de multiples fonctions de d\'ecision. On montre que la variante lin\'eaire de ce cadre admet une interpr\'etation probabilistique simple, tandis que la version plus g\'en\'erale peut s'interpr\'eter en terme de noyaux. Nous proposons une m\'ethode d'optimisation efficace et nous \'evaluons le mod\`ele sur un probl\`eme de reconnaissance de chiffres manuscrits et de classification de textures.
}
\RRabstract{
It is now well established that sparse signal models are well suited to
restoration tasks and can effectively be learned from audio, image,
and video data. Recent research has been aimed at learning {\em
discriminative} sparse models instead of purely {\em reconstructive}
ones. This paper proposes a new step in that direction, with a novel
sparse representation for signals belonging to different classes in
terms of a shared dictionary and multiple class-decision functions. 
The linear variant of the proposed model admits a simple
probabilistic interpretation, while its most general variant 
admits an interpretation in terms of kernels. An 
optimization framework for learning all the components of the proposed
model is presented, along with experimental results on standard
handwritten digit and texture classification tasks.
}
\RRmotcle{parcimonie, sparsit\'e, classification}
\RRkeyword{sparsity, classification}
\RRprojets{Willow}
\RRtheme{\THCog} 
\URRocq 
\RRNo{6652}

\begin{document}
\makeRR   

\section{Introduction}
Sparse and overcomplete image models were first introduced in
\cite{field} for modeling the spatial receptive fields of simple cells
in the human visual system. The linear decomposition of a signal using
a few atoms of a \emph{learned} dictionary, instead of predefined
ones--such as wavelets--has recently led to state-of-the-art results
for numerous low-level image processing tasks such as denoising
\cite{elad}, showing that sparse models are well adapted to natural
images. Unlike principal component analysis decompositions, these models are most ofen overcomplete, with a number of basis elements greater than the
dimension of the data.  Recent research has shown that sparsity helps
to capture higher-order correlation in data: In \cite{huang,wright},
sparse decompositions are used with predefined dictionaries for face
and signal recognition. In \cite{lee2}, dictionaries are learned for a
reconstruction task, and the sparse decompositions are then used a
posteriori within a classifier. In \cite{mairal4}, a
\emph{discriminative} method is introduced for various classification
tasks, learning one dictionary per class; the classification process
itself is based on the corresponding reconstruction error, and does not
exploit the actual decomposition coefficients. In \cite{ranzato3},
a generative model for document representation is learned at the same
time as the parameters of a deep network structure.  The framework we
present in this paper extends these approaches by learning
simultaneously a single shared dictionary as well as multiple decision
functions for different signal classes in a mixed generative and
discriminative formulation (see also \cite{Fernando}, where a different discrimination term is added to the classical reconstructive one for supervised dictionary learning via class supervised simultaneous orthogonal matching pursuit).. Similar joint generative/discriminative
frameworks have started to appear in probabilistic approaches to
learning, e.g., \cite{blei,holub,lasserre2,ng2,hinton,winn}, but not, to
the best of our knowledge, in the sparse dictionary learning
framework. Section \ref{sec:sdl} presents the formulation and Section
\ref{sec:prob} its interpretation in term of probability and kernel frameworks. The optimization procedure is
detailed in Section \ref{sec:optimization}, and experimental results
are presented in Section \ref{sec:experiments}.

\section{Supervised dictionary learning}
\label{sec:sdl}
We present in this section the core of the proposed model. We
start by describing how to perform sparse coding in a supervised
fashion,  then show how to simultaneously learn a
discriminative/reconstructive dictionary and a classifier.
 
\subsection{Supervised Sparse Coding}
In classical {\em sparse coding} tasks, one considers a signal $\x$ in
$\Real^n$ and a {\em fixed} dictionary $\D=[\d_1,\ldots,\d_k]$ in
$\Real^{n\times k}$ (allowing $k>n$, making the dictionary
overcomplete). In this setting, sparse coding with an $\ell_1$
regularization\footnote{
The $\ell_p$ regularization term of a vector $\x$ for
$p \geq 0$ is defined as $||\x||_p^p=(\sum_{i=1}^n
|\x[i]|^p)$. $||.||_p$ is a norm when $p \geq 1$. When $p=0$,
it counts the number of non-zeros elements in the vector.}  amounts to
computing
\begin{equation}
\R^\star(\x,\D)  =  \min_{\alphab \in \Real^k} ||\x-\D\alphab||_2^2 + \lambda_1 ||\alphab||_1. \label{eq:recon}
\end{equation}
It is well known in the statistics, optimization, and compressed
sensing communities that the $\ell_1$ penalty yields a sparse
solution, very few non-zero coefficients in $\alphab$, ~\cite{donoho3},
although there is no explicit analytic link between the value of
$\lambda_1$ and the effective sparsity that this model yields.  Other sparsity
penalties using the $\ell_0$ (or more generally $\ell_p$) regularization
can be used as well. Since it uses a proper norm, the $\ell_1$
formulation of sparse coding is a convex problem, which makes the
optimization tractable with algorithms such as those introduced in
\cite{efron,hale}, and has proven in our proposed framework to be more stable than its $\ell_0$
counterpart, in the sense that the resulting decompositions are less
sensitive to small perturbations of the input signal $\x$. Note that
sparse coding with an $\ell_0$ penalty is an NP-hard problem and is
often approximated using greedy algorithms.

In this paper, we consider a different setting, where the signal may
belong to any of $p$ different classes. We model the signal $\x$ using a
single shared dictionary $\D$ and a set of $p$ decision functions
$g_i(\x,\alphab,\thetab)$ ($i=1,\ldots,p$) acting on  $\x$
and its sparse code $\alphab$ over $\D$.  The function $g_i$ should be
positive for any signal in class $i$ and negative otherwise. The
vector $\thetab$ parametrizes the model and will be {\em jointly} learned
with $\D$. In the following, we will consider two kinds of decision
functions:\\
{\bf (i) linear in $\alphab$}:
$g_i(\x,\alphab,\thetab) = \w_i^T\alphab+b_i$, where $\thetab= \{\w_i
\in \Real^k, b_i \in \Real\}_{i=1}^p$, and the vectors $\w_i$
($i=1,\ldots,p$) can be thought of as $p$ linear models for the
coefficients $\alphab$, with the scalars $b_i$ acting as biases; \\ 
{\bf (ii) bilinear in $\x$ and $\alphab$}:
$g_i(\x,\alphab,\thetab) = \x^T\W_i\alphab+b_i$, where $\thetab =
\{\W_i \in \Real^{n \times k}, b_i \in \Real\}_{i=1}^p$. Note that the
number of parameters in (ii) is greater than in (i), which allows for
richer models.  One can interpret $\W_i$ as a filter encoding the
input signal $\x$ into a model for the coefficients $\alphab$, which
has a role similar to the encoder in \cite{ranzato} but for a
discriminative task.

Let us define {\em softmax} discriminative cost functions as
$$\C_i(x_1,...,x_p) = \log(\sum_{j=1}^p e^{x_j-x_i})$$ for
$i=1,\ldots,p$. These are multiclass versions of the logistic
function, enjoying properties similar to that of the hinge loss from
the SVM literature, while being differentiable. Given some input
signal $\x$ and fixed (for now) dictionary $\D$ and parameters
$\thetab$, the {\em supervised sparse coding} problem for the class
$p$ can be defined as computing
\begin{equation}
\S_i^\star(\x,\D,\thetab) = \min_{\alphab} \S_i(\alphab,\x,\D,\thetab),
\label{eq:convex}
\end{equation}
where
\begin{equation}
\S_i(\alphab,\x,\D,\thetab)=\C_i(\{g_j(\x,\alphab,\thetab)\}_{j=1}^p) + \lambda_0 ||\x-\D\alphab||_2^2
+ \lambda_1 ||\alphab||_1. \label{eq:ssc1}
\end{equation}
Note the explicit incorporation of the classification and discriminative component into sparse coding, in addition to the classical reconstructive term (see \cite{Fernando} for a different classificaiton component). In turn, any solution to this problem provides a straightforward
classification procedure, namely:
\begin{equation}
i^\star(\x,\D,\thetab)= \argmin_{i=1,\ldots,p} \S_i^\star(\x,\D,\thetab).\label{eq:class}
\end{equation}

Compared with earlier work using one dictionary per class \cite{mairal4}, this
model has the advantage of letting multiple classes share some features, and
uses the coefficients $\alphab$ of the sparse representations as part of the
classification procedure, thereby following the works from \cite{huang,lee2,wright}, 
but with learned representations optimized for the classification task similar to \cite{blei,Fernando}.
As shown in Section 3, this formulation has a straightforward
probabilistic interpretation, but let us first see how to learn
the dictionary $\D$ and the parameters $\thetab$ from training data.

\subsection{SDL: Supervised Dictionary Learning}

Let us assume that we are given $p$ sets of training data $T_i$,
$i=1,\dots,p$, such that all samples in $T_i$ belong to class
$i$. 
The most direct method for learning $\D$ and $\thetab$ is to minimize with respect to these variables the mean value of $\S_i^\star$,
with an $\ell_2$ regularization term to prevent overfitting:
\begin{equation}
\min_{\D,\thetab} \Big(\sum_{i=1}^p \sum_{j \in T_i}
\S_i^\star(\x_j,\D,\thetab) \Big) + \lambda_2 ||\thetab||_2^2, ~~\text{s.t.}~~ \forall ~ i = 1,\ldots,k,~~||\d_i||_2 \leq 1.\label{eq:generative}
\end{equation}
Since the reconstruction errors $||\x-\D\alphab||_2^2$ are invariant
to scaling simultaneously $\D$ by a scalar and $\alphab$ by its
inverse, constraining the $\ell_2$ norm of columns of $\D$ prevents
any transfer of energy between these two variables, which would have
the effect of overcoming the sparsity penalty. Such a constraint is
classical in sparse coding \cite{elad}. We will refer later to this model as
\textsf{SDL-G} (supervised dictionary learning, generative).

Nevertheless, since the classification procedure from Eq.~(\ref{eq:class}) will compare the different residuals $\S_i^\star$ of a given signal for $i=1,\ldots,p$, a more discriminative approach is to not only make the
 $\S_i^\star$ small for signals with label $i$, as in (\ref{eq:generative}), but also make the value
 of $\S_j^\star$ greater than $\S_i^\star$ for $j$ different
 than $i$, which is the purpose of the softmax function $\C_i$. This leads to:
 \begin{equation}
 \min_{\D,\thetab} \Big(\sum_{i=1}^p\sum_{j \in T_i} \C_i(\{\S_l^\star(\x_j,\D,\thetab)\}_{l=1}^p)\Big) + \lambda_2 ||\thetab||_2^2 ~~\text{s.t.}~~ \forall ~ i = 1,\ldots,k,~~||\d_i||_2 \leq 1. \label{eq:discr}
 \end{equation}
 As detailed below, this problem is more difficult to solve than Eq.~(\ref{eq:generative}), and
 therefore we adopt instead a mixed
 formulation between the minimization of the generative Eq.~(\ref{eq:generative}) and its discriminative
 version (\ref{eq:discr}),~\cite{ng2}---that is,
 \begin{eqnarray}
& &\!\!\! \min_{\D,\thetab} \Big(\sum_{i=1}^p\sum_{j \in T_i}
 \mu\C_i(\{\S_l^\star(\x_j,\D,\thetab)\}_{l=1}^p)
 + (1-\mu) S_i^\star(\x_j,\D,\thetab) \Big) + \lambda_2 ||\thetab||_2^2 \nonumber \\
&& ~~~~~~~~~~\text{s.t.}~~ \forall i,~~||\d_i||_2 \leq 1,
\label{eq:discr2}
 \end{eqnarray}
where $\mu$ controls the trade-off between reconstruction
from Eq.~(\ref{eq:generative}) and discrimination from
Eq.~(\ref{eq:discr}).  This is the proposed generative/discriminative
model for sparse signal representation and classification from learned
dictionary $\D$ and model $\thetab$. We will refer to this mixed model as
\textsf{SDL-D}, (supervised dictionary learning, discriminative).

Before presenting the proposed
optimization procedure, we provide below two interpretations of the
linear and bilinear versions of our formulation in terms of a
probabilistic graphical model and a kernel.

\section{Interpreting the model\label{sec:prob}}
\subsection{A probabilistic interpretation of the linear model}

Let us first construct a graphical model which gives a probabilistic
interpretation to the training and classification criteria given
above when using a linear model with zero bias (no constant term) on the
coefficients---that is, $g_i(\x,\alphab,\thetab) = \w_i^T\alphab$.  This model
consists of the following components (Figure \ref{fig:graph_model2}):
\begin{figure}
\centering
\scalebox{0.8}{\input{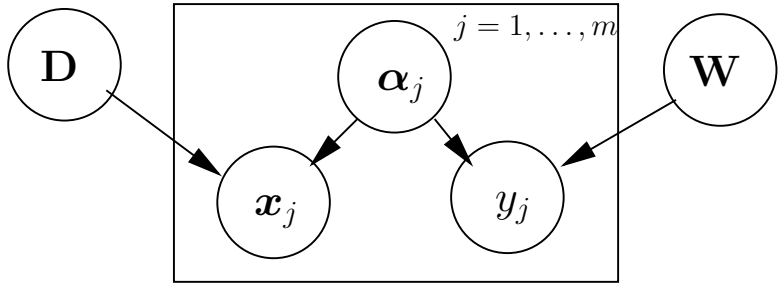}}
\caption{Graphical model for the proposed generative/discriminative learning framework.}\label{fig:graph_model2}
\end{figure}

\noindent $\bullet$ The matrices $\D$ and $\W$ are parameters of the problem, with
a Gaussian prior on $\W$, $p(\W) \propto
e^{-\lambda_2||\W||_2^2}$, and on the columns of $\D$,
$p(\D) \propto \prod_{l=1}^k e^{-\gamma_l||\d_l||_2^2}$, where the $\gamma_l$'s are the Gaussian parameters. All the $\d_l$'s are considered independent of each other. \\
\noindent $\bullet$ The coefficients $\alphab_j$ are latent variables with a Laplace
prior, $p(\alphab_j) \propto e^{-\lambda_1||\alphab_j||_1}$.\\
\noindent $\bullet$ The signals $\x_j$ are generated according to a Gaussian probability
distribution conditioned on $\D$ and $\alphab_j$,
$p(\x_j | \alphab_j,\D) \propto e^{-\lambda_0||\x_j-\D\alphab_j||_2^2}$.
All the $\x_j$'s are considered independent from each other.\\
\noindent $\bullet$ The labels $y_j$ are generated according to a probability
distribution conditioned on $\W$ and $\alphab_j$, and given by $p(y_j=i
| \alphab_j,\W) = {e^{-\W_{i}^T\alphab_j}}/{\sum_{l=1}^p
e^{-\W_{l}^T \alphab_j}}$. Given $\D$ and $\W$, all the triplets $(\alphab_j,\x_j,y_j)$ are independent.

What is commonly called ``generative training'' in the
literature~(e.g., \cite{lasserre2,ng2}), amounts to finding the maximum
likelihood for $\D$ and $\W$ according to the joint distribution
$p(\{\x_j,y_j\}_{j=1}^m,\D,\W)$, where the $\x_j$'s and the $y_j$'s
are respectively the training signals and their labels.  It can easily
be shown (details omitted due to space limitations) that there is an
equivalence between this generative training and our formulation in
Eq. (\ref{eq:generative}) under MAP approximations.\footnote{We are
also investigating how to properly estimate $\D$ by marginalizing over
$\alphab$ instead of maximizing with respect to that parameter.}
Although joint generative modeling of $\x$ and $y$ through a shared
representation, e.g., \cite{blei}, has shown great promise, we show in
this paper that a more discriminative approach is desirable.
``Discriminative training'' is slightly different and amounts to
maximizing $p(\{y_j\}_{j=1}^m,\D,\W|\{\x_j\}_{j=1}^m)$ with respect to
$\D$ and $\W$: Given some input data, one finds the best parameters
that will predict the labels of the data. The same kind of MAP
approximation relates this discriminative training formulation to the
discriminative model of Eq. (\ref{eq:discr}) (again, details omitted
due to space limitations). The mixed approach from
Eq. (\ref{eq:discr2}) is a classical trade-off between generative and
discriminative (e.g., \cite{lasserre2,ng2}), where generative components
are often added to discriminative frameworks to add robustness, e.g.,
to noise and occlusions (see examples of this for the model in \cite{Fernando}).

\subsection{A kernel interpretation of the bilinear model}
Our bilinear model with
$g_i(\x,\alphab,\thetab)=\x^T\W_i\alphab+b_i$ does not admit a
straightforward probabilistic interpretation.  On the other hand, it
can easily be interpreted in terms of kernels: Given two signals
$\x_1$ and $\x_2$, with coefficients $\alphab_1$ and $\alphab_2$,
using the kernel $K(\x_1,\x_2)=\alphab_1^T\alphab_2 \x_1^T\x_2$ in a
logistic regression classifier amounts to finding a decision function
of the same form as (ii).  It is a product of two linear kernels, one
on the $\alphab$'s and one on the input signals $\x$.  Interestingly,
Raina et al.~\cite{lee2} learn a dictionary adapted to reconstruction
on a training set, then train an SVM a posteriori on the decomposition
coefficients $\alphab$.  They derive and use a Fisher kernel, which can be
written as $K'(\x_1,\x_2)=\alphab_1^T\alphab_2\r_1^T\r_2$ in this
setting, where the $\r$'s are the residuals of the decompositions.
Experimentally, we have observed that the kernel $K$, where the
signals $\x$ replace the residuals $\r$, generally yields a level of
performance similar to $K'$, and often actually does better when the
number of training samples is small or the data are noisy.

\section{Optimization procedure}
\label{sec:optimization}

Classical dictionary learning techniques (e.g., \cite{aharon,field,lee2}), address the problem of learning a reconstructive dictionary $\D$ in $\Real^{n \times k}$ well adapted to a training set $T$ as
\begin{equation}
\min_{\D,\alphab} \sum_{j \in T} ||\x_j-\D\alphab_j||_2^2+\lambda_1||\alphab_j||_1,
\end{equation}
which is not jointly convex in $(\D,\alphab)$, but convex with respect to each unknown when the other one is fixed. This is why block coordinate descent on $\D$ and $\alphab$ performs reasonably well \cite{aharon,field,lee2}, although not necessarily providing the global optimum. Training when $\mu=0$ (generative case), i.e., from Eq. (\ref{eq:generative}), enjoys similar properties and can be addressed with the same optimization procedure. Equation (\ref{eq:generative}) can be rewritten as:
\begin{equation}
\min_{\D,\thetab,\alphab} \Big(\sum_{i=1}^p \sum_{j \in T_i}
\S_i(\x_j,\alphab_j,\D,\thetab) \Big) + \lambda_2 ||\thetab||_2^2, ~~\text{s.t.}~~ \forall ~ i = 1,\ldots,k,~~||\d_i||_2 \leq 1.
\end{equation}
Block coordinate descent consists therefore of iterating between \emph{supervised sparse coding}, where $\D$ and $\thetab$ are fixed and one optimizes with respect to the $\alphab$'s
and \emph{supervised dictionary update}, where the coefficients $\alphab_j$'s are fixed, but $\D$ and $\thetab$ are updated. Details on how to solve these two problems are given in Section 4.1 and 4.2.

\begin{figure}
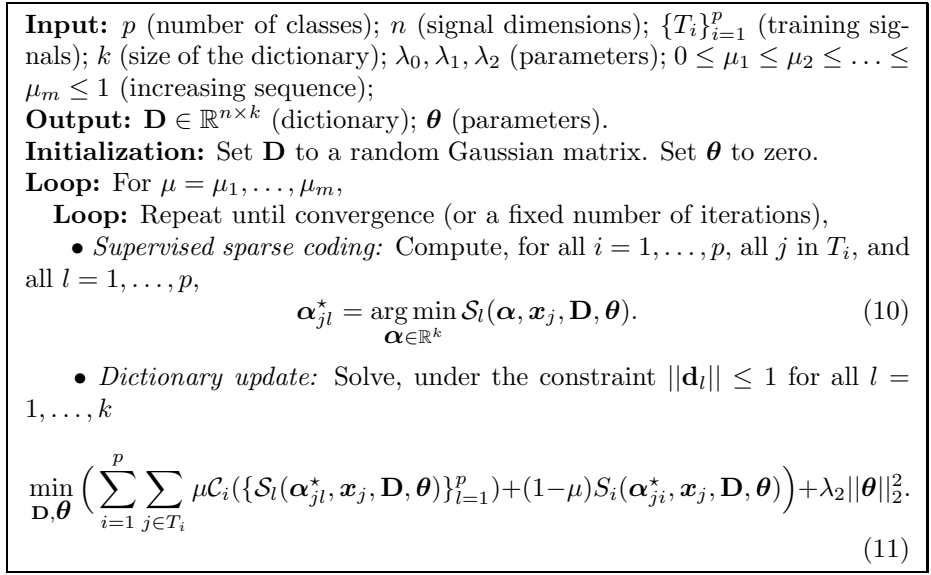

\centering
\fbox{
\begin{minipage}{0.96\linewidth}
\noindent {\bf Input:} $p$ (number of classes); $n$ (signal dimensions); $\{T_i\}_{i=1}^p$ (training signals); $k$ (size of the dictionary); $\lambda_0,\lambda_1,\lambda_2$ (parameters); $0 \leq \mu_1 \le \mu_2 \le \ldots \le \mu_m \leq 1$ (increasing sequence);

\noindent {\bf Output:} $\D \in \Real^{n \times k}$ (dictionary); $\thetab$ (parameters).

\noindent {\bf Initialization:} Set $\D$ to a random Gaussian matrix. Set $\thetab$ to zero.

\noindent {\bf Loop:} For $\mu = \mu_1, \ldots, \mu_m$,\\
\hspace*{0.25cm} {\bf Loop:} Repeat until convergence (or a fixed number of iterations),\\
\hspace*{0.5cm} $\bullet$ \emph{Supervised sparse coding:} Compute, for all $i=1,\ldots,p$, all $j$ in $T_i$, and all $l=1,\ldots,p$,
\begin{equation}
\alphab_{jl}^\star = \argmin_{\alphab \in \Real^k} \S_l(\alphab,\x_j,\D,\thetab).\label{eq:ssc}
\end{equation}
\hspace*{0.5cm} $\bullet$ \emph{Dictionary update:} Solve, under the constraint $||\d_l||\leq 1$ for all $l=1,\ldots,k$ 
\begin{equation}
\min_{\D,\thetab} \Big(\sum_{i=1}^p\sum_{j \in T_i}
 \mu\C_i(\{\S_l(\alphab_{jl}^\star,\x_j,\D,\thetab)\}_{l=1}^p)
 + (1-\mu) S_i(\alphab_{ji}^\star,\x_j,\D,\thetab) \Big) + \lambda_2 ||\thetab||_2^2.\label{eq:dicoUpdate}
\end{equation}
\end{minipage}
}
\caption{{\it SDL:} Supervised dictionary learning algorithm.} \label{fig:algo}
\end{figure}

The discriminative version of SDL from Eq. (\ref{eq:discr}) is more problematic. The minimization of the term $\C_i(\{\S_l(\alphab_{jl},\x_j,\D,\thetab)\}_{l=1}^p)$ with respect to $\D$ and $\thetab$ when the $\alphab_{jl}$'s are fixed, is not convex in general, and does not necessarily decrease the first term of Eq.~(\ref{eq:discr}), i.e., $\C_i(\{\S_l^\star(\x_j,\D,\thetab)\}_{l=1}^p)$. To reach a local minimum for this difficult problem, we have chosen a continuation method, starting from the generative case and ending with the discriminative one as in \cite{mairal4}. The algorithm is presented on Figure \ref{fig:algo}, and details on  the hyperparameters' settings are given in Section 5.

\subsection{Supervised sparse coding}

The supervised sparse coding problem from Eq. (\ref{eq:ssc}) ($\D$ and $\thetab$ are fixed in this step), amounts to minimizing a convex function under an $\ell_1$ penalty. The \emph{fixed-point continuation method} (FPC) from \cite{hale} achieves state-of-the-art results in terms of convergence speed for this class of problems. It has proven in our experiments to be simple, efficient, and well adapted to our supervised sparse coding problem. Algorithmic details are given in \cite{hale}. For our specific problem, denoting by $f$ the convex function to minimize, this method only requires $\nabla f$ and a bound on the spectral norm of its Hessian $\H_f$. Since the we have chosen decision functions $g_i$ in Eq. (\ref{eq:ssc}) which are linear in $\alphab$, there exists, for each signal $\x$ to be sparsely represented, a matrix $\A$ in $\Real^{k \times p}$ and a vector $\b$ in $\Real^p$ such that
\begin{equation}
\left\{
\begin{aligned}
f(\alphab) & = & \C_i(\A^T\alphab+\b) + \lambda_0 ||\x-\D\alphab||_2^2, \nonumber \\
\nabla f(\alphab) & = & \A \nabla \C_i(\A^T\alphab+\b) -2 \lambda_0 \D^T(\x-\D\alphab),
\end{aligned}\right.
\end{equation}
and it can be shown that, if $||\U||_2$ denotes the spectral norm of a matrix $\U$ (which is the magnitude of its largest eigenvalue), then $||\H_f||_2 \leq (1-\frac{1}{p}) ||\A^T\A||_2^2 + 2 \lambda_0 ||\D^T\D||_2$. In the case where $p=2$ (only two classes), we can obtain a tighter bound, $||\H_f(\alphab)||_2 \leq e^{-\C_1(\A^T\alphab)-\C_2(\A^T\alphab)}||\a_2-\a_1||_2^2+2\lambda_0||\D^T\D||_2$, where $\a_1$ and $\a_2$ are the first and second columns of $\A$.

\subsection{Dictionary update}

The problem of updating $\D$ and $\thetab$ in Eq. (\ref{eq:dicoUpdate}) is not convex in general (except when $\mu$ is close to 0), but a local minimum can be obtained using projected gradient descent (as in the general literature on dictionary learning, this local minimum has experimentally been found to be
good enough for our formulation). 
Denoting $E(\D,\thetab)$ the function we want to minimize in Eq. (\ref{eq:dicoUpdate}), we just need the partial derivatives of $E$ with respect to $\D$ and the parameters $\thetab$. Details when using the linear model for the $\alphab$'s, $g_i(\x,\alphab,\thetab) = \w_i^T\alphab+b_i$, and $\thetab=\{\W \in \Real^{k \times p}, \b \in \Real^p\}$, are
\begin{equation}
\left\{
\begin{aligned}
\frac{\partial E}{\partial \D}& =& -2\lambda_0 \big( \sum_{i=1}^p\sum_{j \in T_i}\sum_{l=1}^p \omega_{jl}(\x_j-\D\alphab_{jl}^\star) \alphab_{jl}^{\star T}\big),\\
\frac{\partial E}{\partial \W}& =&  \sum_{i=1}^p\sum_{j \in T_i}\sum_{l=1}^p \omega_{jl} \alphab_{jl}^\star \nabla\C_l^T(\W^T\alphab_{jl}^\star+\b), \\
\frac{\partial E}{\partial \b}& =&  \sum_{i=1}^p\sum_{j \in T_i}\sum_{l=1}^p \omega_{jl} \nabla\C_l(\W^T\alphab_{jl}^\star+\b), 
\end{aligned}\right.
\end{equation}
where
\begin{equation}
\omega_{jl} = \mu \nabla \C_i( \{\S_m(\alphab_{jm}^\star,\x_j,\D,\thetab)\}_{m=1}^p )[l] +(1-\mu){\mathbf 1}_{l = i}.
\end{equation}
Partial derivatives when using our model with the bilinear decision functions
$g_i(\x,\alphab,\theta)=\x^T\W_i\alphab+b_i$ are not given in this paper because of space limitations.

\section{Experimental validation}
\label{sec:experiments}
We compare in this section a reconstructive approach, dubbed
\textsf{REC}, which consists of learning a reconstructive dictionary
$\D$ as in \cite{lee2} and then learning the parameters $\thetab$ a
posteriori; SDL with generative training (dubbed \textsf{SDL-G}); and
SDL with discriminative learning (dubbed \textsf{SDL-D}). We also
compare the performance of the linear (\textsf{L}) and bilinear
(\textsf{BL}) decision functions.

Before presenting experimental results, let us briefly discuss the
choice of the five model parameters $\lambda_0$, $\lambda_1$,
$\lambda_2$, $\mu$ and $k$ (size of the dictionary). Tuning all of
them using cross-validation is cumbersome and unnecessary since some
simple choices can be made, some of which can be done sequentially. We
define first the \emph{sparsity} parameter
$\kappa=\frac{\lambda_1}{\lambda_0}$, which dictates how sparse the
decompositions are. When the input data points have unit $\ell_2$
norm, choosing $\kappa=0.15$ was empirically found to be a good
choice. The number of parameters to learn is linear in $k$, the number
of elements in the dictionary $\D$. For reconstructive tasks, $k=256$
is a typical value often used in the literature (e.g.,
\cite{aharon}). Nevertheless, for discriminative tasks, increasing the
number of parameters is likely to allow overfitting, and smaller
values like $k=64$ or $k=32$ are preferred.  The scalar $\lambda_2$ is
a regularization parameter for preventing the model to overfit the
input data. As in logistic regression or support vector machines, this
parameter is crucial when the number of training samples is
small. Performing cross validation with the fast method \textsf{REC}
quickly provides a reasonable value for this parameter, which can be
used afterward for \textsf{SDL-G} or \textsf{SDL-D}.

Once $\kappa$, $k$ and $\lambda_2$ are chosen, let us see how to find $\lambda_0$. In logistic regression, a projection matrix maps input data onto a softmax function, and its shape and scale are adapted so that it becomes discriminative according to an underlying probabilistic model. In the model we are proposing, the functions $\S_i^\star$ are also mapped onto a softmax function, and the parameters $\D$ and $\thetab$ are adapted (learned) in such a way that $\S_i^\star$ becomes discriminative. However, for a fixed $\kappa$, the second and third terms of $\S_i^\star$, namely $\lambda_0||\x-\D\alphab||_2^2$ and $\lambda_0\kappa||\alphab||_1$, are not freely scalable when adapting $\D$ and $\thetab$, since their magnitudes are bounded. 
$\lambda_0$ plays the important role of controlling the trade-off between reconstruction and discrimination in Eq. (\ref{eq:ssc1}). First, we perform cross-validation for a few iterations with $\mu=0$ to find a good value for \textsf{SDL-G}. Then, a scale factor making the $\S_i^\star$'s discriminative for $\mu > 0$ can be chosen during the optimization process: Given a set of $\S_i^\star$'s, one can compute a scale factor $\gamma$ such that $\gamma = \argmin_\gamma \sum_{i=1}^p \sum_{j \in T_i} \C_i(\{\gamma \S_l^\star(\x_j,\D,\W)\})$. We therefore propose the following strategy, which has proven to be efficient during our experiments: Starting from small values for $\lambda_0$ and a fixed $\kappa$, we apply the algorithm in Figure \ref{fig:algo}, and after a supervised sparse coding step, we compute the best scale factor $\gamma$, and replace $\lambda_0$ and $\lambda_1$ by $\gamma \lambda_0$ and $\gamma \lambda_1$. Typically, applying this procedure during the first $10$ iterations has proven to lead to reasonable values for this parameter.

Since we are following a continuation path starting from $\mu=0$ to $\mu=1$, the optimal value of $\mu$ is found along the path by measuring the classification performance of the model on a validation set during the optimization.

\subsection{Digits recognition}
In this section, we present experiments on the popular MNIST \cite{lecun} and USPS handwritten digit datasets. MNIST is composed of $70\,000$ images of $28 \times 28$ pixels, $60\,000$ for training, $10\,000$ for testing, each of them containing a handwritten digit. USPS is composed of 7291 training images and 2007 test images.  As it is often done in classification, we have chosen to learn pairwise binary classifiers, one for each pair of digits. Although we have presented a multiclass framework, pairwise binary classifiers have proven to offer a slightly better performance in practice. Five-fold cross validation has been performed to find the best pair $(k,\kappa)$. The tested values for $k$ are $\{24,32,48,64,96\}$, and for $\kappa$, $\{0.13, 0.14, 0.15, 0.16, 0.17\}$. Then, we have kept the three best pairs of parameters and used them to train three sets of pairwise classifiers. For a given patch $\x$, the test procedure consists of selecting the class which receives the most votes from the pairwise classifiers. All the other parameters are obtained using the procedure explained above.
Classification results are presented on Table \ref{table:mnist} when using the linear model. We see that for the linear model \textsf{L}, \textsf{SDL-D L} performs the best. \textsf{REC BL} offers a larger feature space and performs better than \textsf{REC L}. Nevertheless, we have observed no gain by using \textsf{SDL-G BL} or \textsf{SDL-D BL} instead of \textsf{REC BL}. Since the linear model is already performing very well, one side effect of using \textsf{BL} instead of \textsf{L} is to increase the number of free parameters and thus to cause overfitting. 
Note that the best error rates published on these datasets (without any modification of the training set) are $0.60\%$ \cite{ranzato} for MNIST and $2.4\%$ \cite{haasdonk} for USPS, 
using methods tailored to these tasks, whereas ours is generic and has not been tuned to  the handwritten digit classification domain.

\begin{table}
\centering
\footnotesize{
\centering
\begin{tabular}{*{8}{|c|}}
\hline
      & \textsf{REC L} & \textsf{SDL-G L} & \textsf{SDL-D L} & \textsf{REC BL} & k-\textsf{NN}, $\ell_2$ &  SVM-Gauss  \\
\hline
MNIST &  4.33          &   3.56           &  {\bf 1.05}      & 3.41            &     5.0                &   1.4  \\
\hline
USPS  &  6.83          &   6.67           &  {\bf 3.54}      & 4.38            &     5.2                 &   4.2   \\
\hline
\end{tabular} 
}
\caption{Error rates on MNIST and USPS datasets in percents from the \textsf{REC}, \textsf{SDL-G L} and \textsf{SDL-D L} approaches, 
compared with k-nearest neighbor and SVM with a Gaussian kernel \cite{lecun}.}\label{table:mnist}
\end{table}

The purpose of our second experiment is not to measure the raw performance of our algorithm, but to answer the question \emph{``are the obtained dictionaries $\D$ discriminative per se or is the pair ($\D$,$\thetab$) discriminative?''}. To do so, we have trained on the USPS dataset $10$ binary classifiers, one per digit in a one vs all fashion on the training set. For a given value of $\mu$, we obtain $10$ dictionaries $\D$ and $10$ sets of parameters $\thetab$, learned by the \textsf{SDL-D L} model.

To evaluate the discriminative power of the dictionaries $\D$, we discard the learned parameters $\thetab$ and use the dictionaries as if they had been learned in a reconstructive \textsf{REC} model: For each dictionary, we decompose each image from the training set by solving the simple sparse reconstruction problem from Eq.~(\ref{eq:recon}) instead of using supervised sparse coding. This provides us with some coefficients $\alphab$, which we use as features in a linear SVM. Repeating the sparse decomposition procedure on the test set permits us to evaluate the performance of these learned linear SVM.
We plot the average error rate of these classifiers on Figure \ref{fig:curve} for each value of $\mu$. We see that using the dictionaries obtained with discrimative learning ($\mu > 0$, \textsf{SDL-D L}) dramatically improves the performance of the basic linear classifier learned a posteriori on the $\alphab$'s, showing that our learned dictionaries are discriminative per se. Figure \ref{fig:dico9} shows a dictionary adapted to the reconstruction of the MNIST dataset and a discriminative one, adapted to ``$9$ vs all''. 

\begin{figure}[hbtp]
   \centering
   \begin{minipage}{0.7\linewidth}
\psset{yunit=1.5,xunit=7.5}
\begin{pspicture}(0,0)(1.05,3.07)
\psaxes[Dx=0.2,Dy=0.5,ticksize=1pt,labelFontSize=\small]{->}(0,0)(0,0)(1.05,3.07)
\psgrid[gridlabels=0pt](0,0)(1,3)
\fileplot[linecolor=red,linewidth=0.75pt]{lin.data}
\end{pspicture}
\end{minipage}
\caption{Average error rate in percents obtained by our dictionaries learned in a discriminative framework (\textsf{SDL-D L}) for various values of $\mu$, when used in used at test time in a reconstructive framework (\textsf{REC-L}). See text for details.}
\label{fig:curve}
\end{figure}
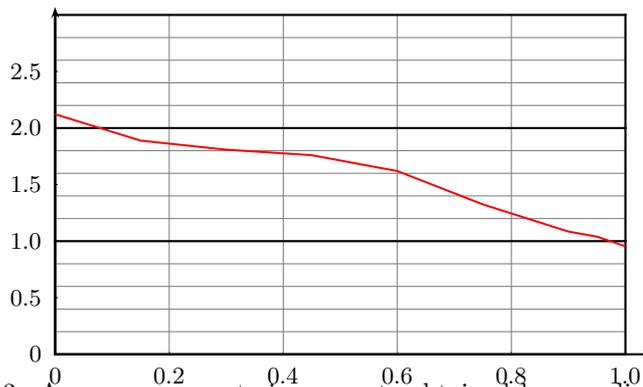
\begin{figure}[hbtp]
   \centering
\subfigure[\textsf{REC, MNIST}]{\includegraphics[width=6.0cm]{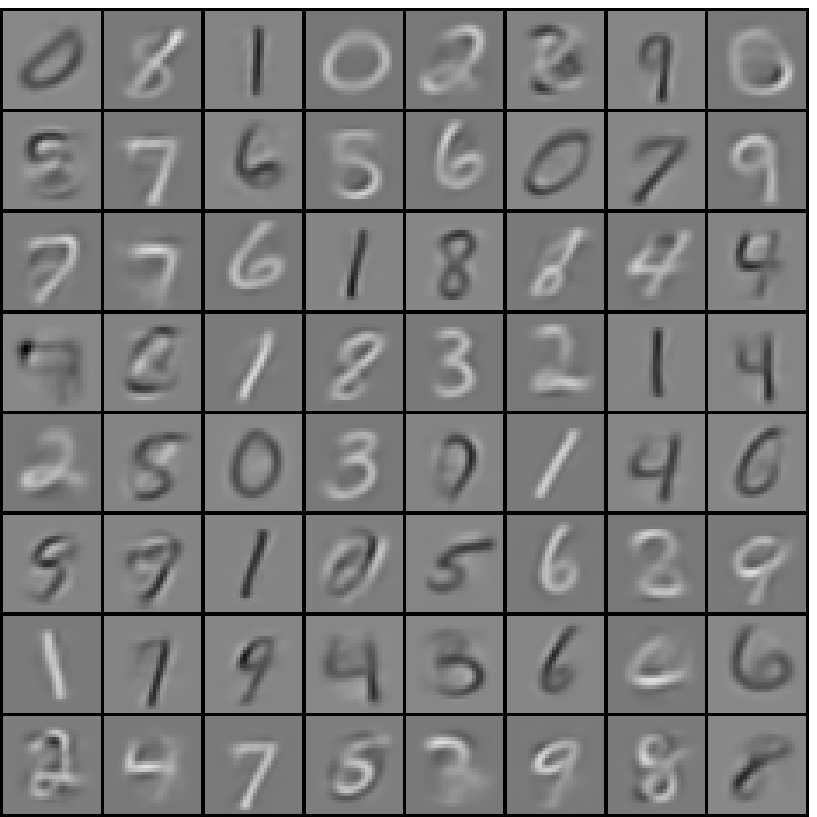}} \hfill
\subfigure[\textsf{SDL-D, MNIST}]{\includegraphics[width=6.0cm]{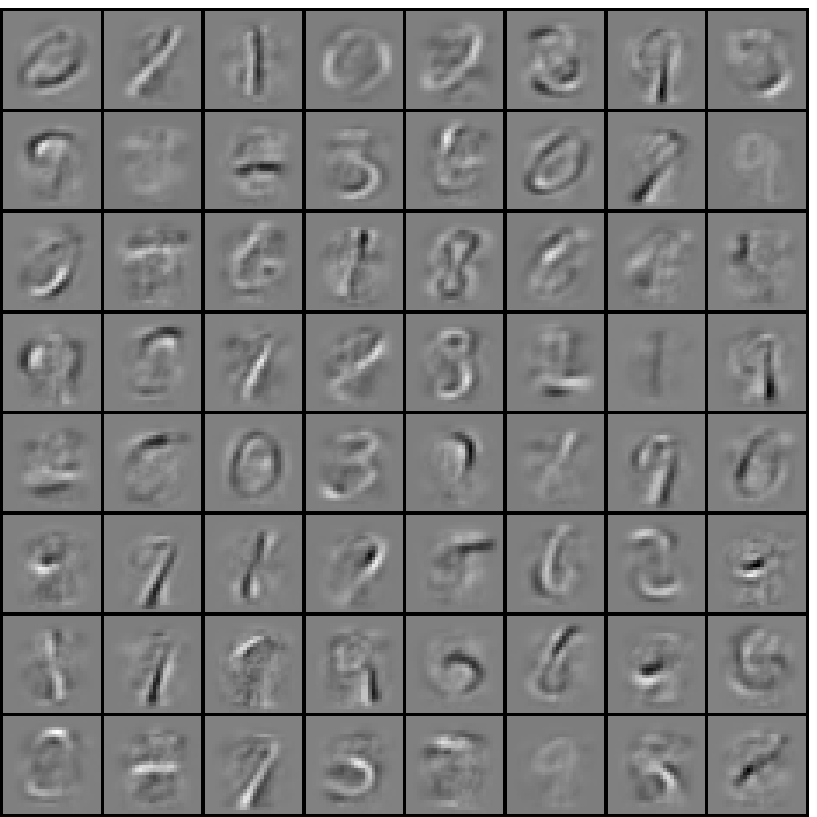}} 
\caption{On the left, a reconstructive dictionary, on the right a discriminative one for the task ``$9$ vs all''.}
\label{fig:dico9}
\end{figure}


\subsection{Texture classification}

\begin{figure}[hbtp]
\subfigure[Texture 1]{\includegraphics[width=6.0cm]{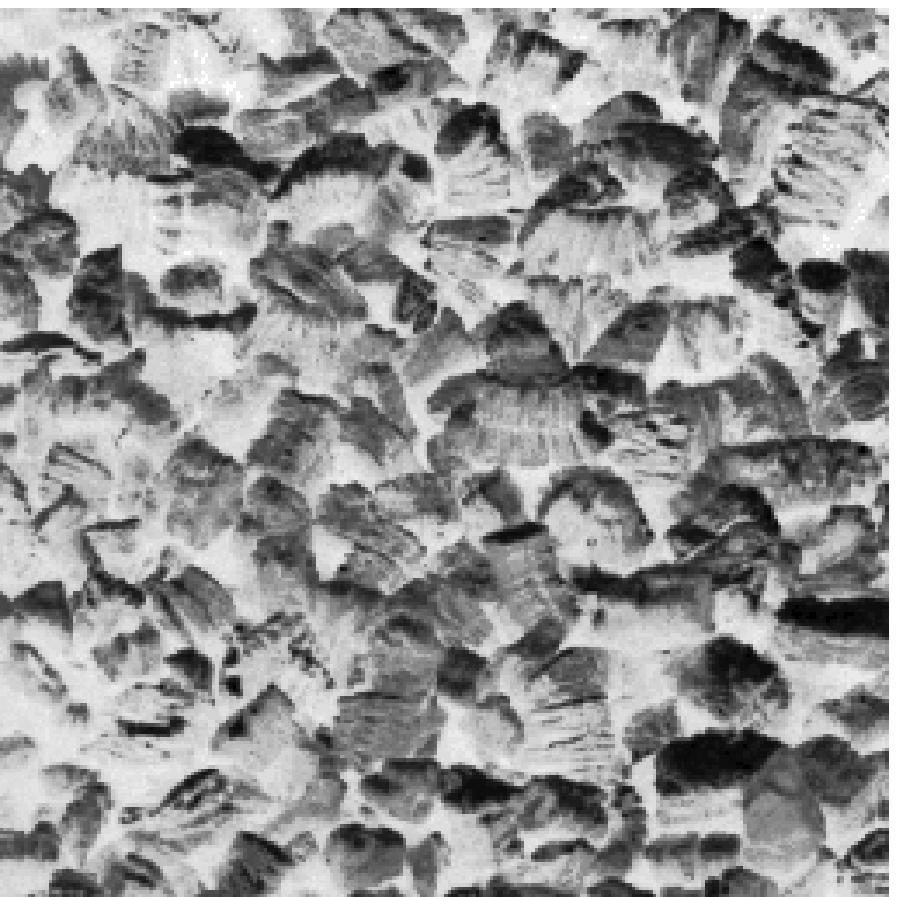}} \hfill
\subfigure[Texture 2]{\includegraphics[width=6.0cm]{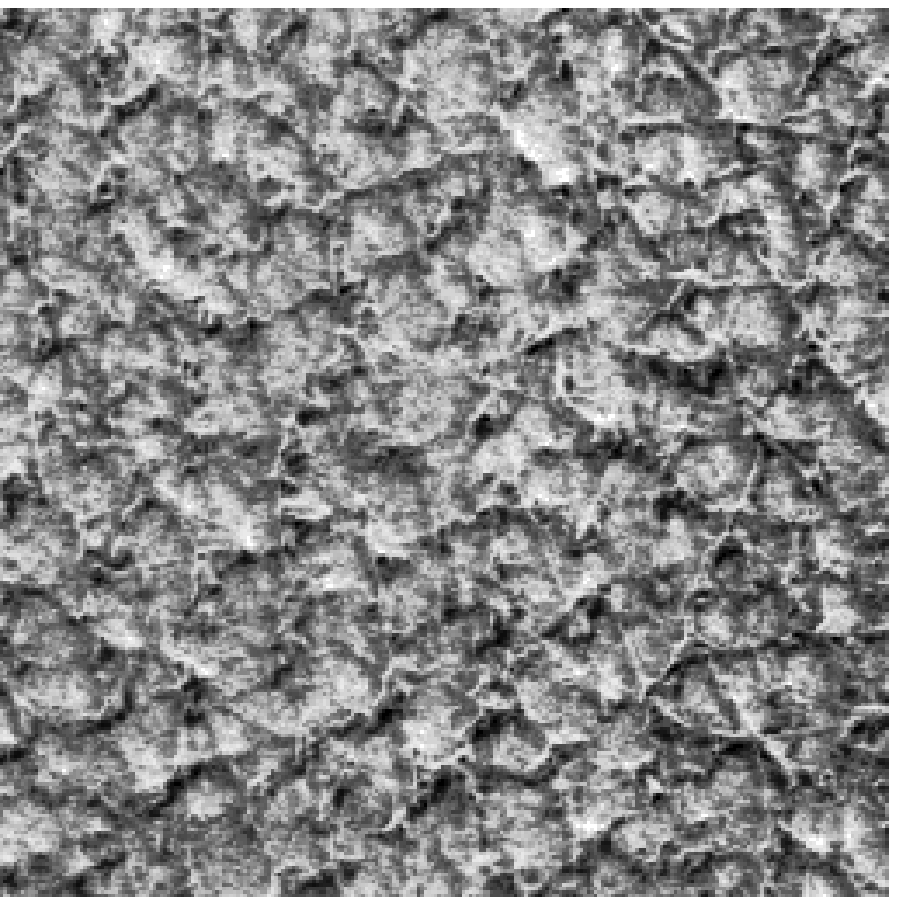}} \hfill
\subfigure[\textsf{REC}]{\includegraphics[width=6.0cm]{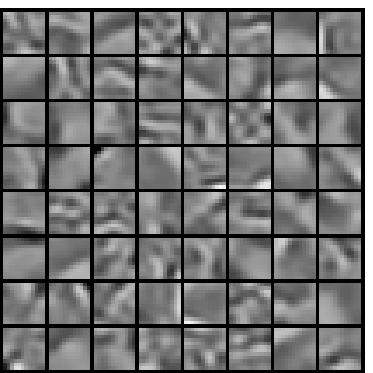}} \hfill
\subfigure[\textsf{SDL-D BL}]{\includegraphics[width=6.0cm]{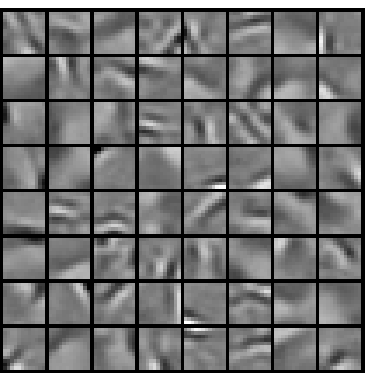}} \hfill
\caption{Top: Test textures. Bottom left: reconstructive dictionary. Bottom right: discriminative dictionary.} \label{fig:tex}
\end{figure}

In the digit recognition task, our \textsf{BL} bilinear framework did not perform better than \textsf{L} and we believe that one of the main reasons is due to the simplicity of the task, where a linear model is rich enough. The purpose of our next experiment is to answer the question \emph{``When is \textsf{BL} worth using?''}. We have chosen to consider two texture images from the Brodatz dataset, presented in Figure \ref{fig:tex}, and to build two classes, composed of $12 \times 12$ patches taken from these two textures. We have compared the classification performance of all our methods, including \textsf{BL}, for a dictionary of size $k=64$ and $\kappa=0.15$. The training set was composed of patches from the left half of each texture and the test sets of patches from the right half, so that there is no overlap between them in the training and test set. Error rates are reported for varying sizes of the training set. This experiment shows that in some cases, the linear model completely fails and \textsf{BL} is necessary. Discrimination helps especially when the size of the training set is particularly valuable for large training sets. Note that we did not perform any cross-validation to optimize the parameters $k$ and $\kappa$ for this experiment. Dictionaries obtained with \textsf{REC} and \textsf{SDL-D BL} are presented in Figure \ref{fig:tex}. Note that though they are visually quite similar, they lead to very different performance.

\begin{table}[t]
\centering
\footnotesize{
\begin{tabular}{|c||c|c|c||c|c|c|c|}
\hline
M & \textsf{REC L} & \textsf{SDL-G L} & \textsf{SDL-D L} & \textsf{REC BL} & \textsf{SDL-G BL} & \textsf{SDL-D BL} & Gain\\
\hline
300 & 48.84 & 47.34 & 44.84  & 26.34  & 26.34  & 26.34 & 0\%         \\
1500 & 46.8 & 46.3 & 42  & 22.7   & 22.3   & 22.3  & 2\%  \\
3000 & 45.17 & 45.1 & 40.6   & 21.99  & 21.22  & 21.22 & 4\% \\
6000 & 45.71 & 43.68 & 39.77 & 19.77  & 18.75  & 18.61 & 6\%   \\
15000 & 47.54 & 46.15 & 38.99  & 18.2   & 17.26 & 15.48 & 15\%   \\
30000 & 47.28  & 45.1  & 38.3  & 18.99 & 16.84  & 14.26 & 25\%  \\
\hline
\end{tabular}
}
\caption{Error rates for the texture classification task using various frameworks and sizes $M$ of training set. The last column indicates the gain between the error rate of \textsf{REC BL} and \textsf{SDL-D BL}.}
\label{table:brodatz}
\end{table}

\section{Conclusion}

We have introduced in this paper a discriminative approach to
supervised dictionary learning that effectively exploits the
corresponding sparse signal decompositions in image classification
tasks, and affords an effective method for learning a shared
dictionary and multiple (linear or bilinear) decision
functions. Future work will be devoted to adapting the proposed
framework to shift-invariant models that are standard in image
processing tasks, but not readily generalized to the sparse dictionary
learning setting. We are also investigating extensions to
unsupervised and semi-supervised learning and applications into natural image classification.


\small{
\bibliographystyle{plain}
\bibliography{IEEEabrv,main}
}

\end{document}